%% file: main.tex
\documentclass[letterpaper]{article} % DO NOT CHANGE THIS
\usepackage{aaai25}  % DO NOT CHANGE THIS
\usepackage{times}  % DO NOT CHANGE THIS
\usepackage{helvet}  % DO NOT CHANGE THIS
\usepackage{courier}  % DO NOT CHANGE THIS
\usepackage[hyphens]{url}  % DO NOT CHANGE THIS
\usepackage{graphicx} % DO NOT CHANGE THIS
\usepackage{natbib}  % DO NOT CHANGE THIS AND DO NOT ADD ANY OPTIONS TO IT
\usepackage{caption} % DO NOT CHANGE THIS AND DO NOT ADD ANY OPTIONS TO IT
\usepackage{algorithm}
\usepackage{algorithmic}

\usepackage{booktabs}
\usepackage{multirow}
\usepackage{tabularx}
\usepackage{subcaption}
\usepackage{amsmath}
\usepackage{bbding}

\usepackage{newfloat}
\usepackage{listings}
\DeclareCaptionStyle{ruled}{labelfont=normalfont,labelsep=colon,strut=off} % DO NOT CHANGE THIS
\floatstyle{ruled}
\newfloat{listing}{tb}{lst}{}
\floatname{listing}{Listing}
\title{Learning with Open-world Noisy Data via Class-independent Margin in Dual Representation Space}
\author {
    Linchao Pan\textsuperscript{\rm 1},
    Can Gao\textsuperscript{\rm 1, \rm 2}\thanks{Corresponding author.},
    Jie Zhou\textsuperscript{\rm 2, \rm 3},
    Jinbao Wang\textsuperscript{\rm 2, \rm 3}
}
\affiliations {
    \textsuperscript{\rm 1}College of Computer Science and Software Engineering, Shenzhen University\\
    \textsuperscript{\rm 2}Guangdong Provincial Key Laboratory of Intelligent Information Processing\\
    \textsuperscript{\rm 3}National Engineering Laboratory for Big Data System Computing Technology, Shenzhen University\\
    linchaopan2022@163.com, 2005gaocan@163.com, jie\_jpu@163.com, wangjb@szu.edu.cn
}

\begin{document}

\maketitle

\begin{abstract}
Learning with Noisy Labels (LNL) aims to improve the model generalization when facing data with noisy labels, and existing methods generally assume that noisy labels come from known classes, called closed-set noise. However, in real-world scenarios, noisy labels from similar unknown classes, i.e., open-set noise, may occur during the training and inference stage. Such open-world noisy labels may significantly impact the performance of LNL methods. In this study, we propose a novel dual-space joint learning method to robustly handle the open-world noise. To mitigate model overfitting on closed-set and open-set noises, a dual representation space is constructed by two networks. One is a projection network that learns shared representations in the prototype space, while the other is a One-Vs-All (OVA) network that makes predictions using unique semantic representations in the class-independent space. Then, bi-level contrastive learning and consistency regularization are introduced in two spaces to enhance the detection capability for data with unknown classes. To benefit from the memorization effects across different types of samples, class-independent margin criteria are designed for sample identification, which selects clean samples, weights closed-set noise, and filters open-set noise effectively. Extensive experiments demonstrate that our method outperforms the state-of-the-art methods and achieves an average accuracy improvement of 4.55\% and an AUROC improvement of 6.17\% on CIFAR80N.
\end{abstract}

\begin{links}
    \link{Code}{https://github.com/iCAN-SZU/LOND-DRS}
\end{links}


\section{Introduction}
Deep Neural Networks (DNNs) have achieved great success in many fields, while their effectiveness heavily relies on a large amount of data with accurate and complete labels. Nevertheless, these high-quality labels are generally annotated by domain experts at the expense of high cost, thus some less-costly techniques are used to collect large-scale datasets, such as web crawling and crowdsourcing \cite{song2022learning}. As a result, these non-expert manners inevitably introduce annotation errors and generate datasets with noisy labels. It has been shown that the over-parameterized DNNs can easily memorize the noisy labels \cite{zhang2017understanding}, thereby degrading model generalization and performance. Therefore, it is crucial to develop effective methods for DNNs to learn from data with noisy labels.

\begin{figure}[t]
    \centering
    \begin{subfigure}{0.48\columnwidth}
        \includegraphics[width=\linewidth]{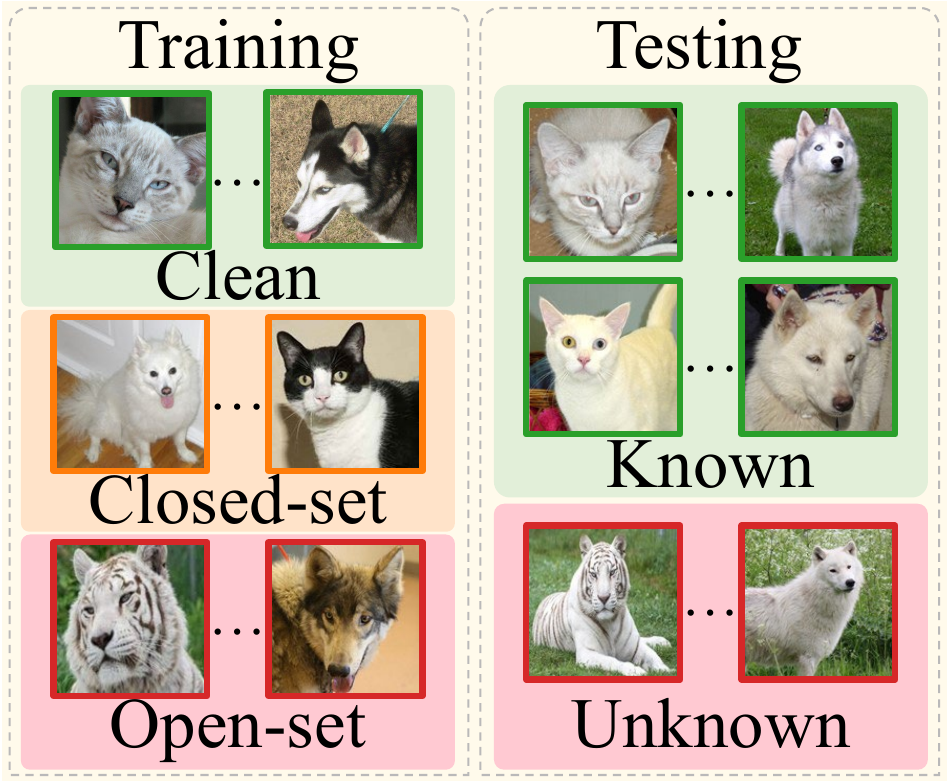}
    \end{subfigure}\hfill
    \begin{subfigure}{0.48\columnwidth}
        \includegraphics[width=\linewidth]{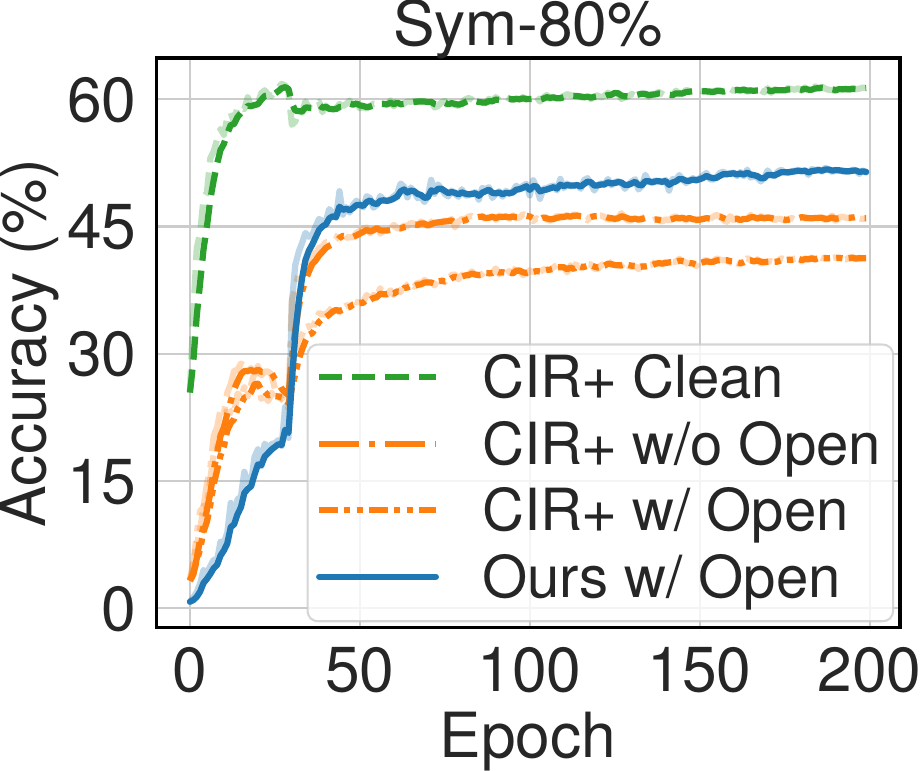}
    \end{subfigure}
    \caption{Illustration of the LOND setup and the effect of open-set noise. \textbf{Left:} In addition to closed-set noise, open-set noise is also present in the training and testing stage. \textbf{Right:} Open-set noise (w/ Open) significantly degrades performance on CIFAR80N with 80\% symmetric noise.} \label{fig:setup}
\end{figure}

Learning with Noisy Labels (LNL) has emerged as an effective learning paradigm for data with noisy labels \cite{song2022learning}. Recently, a variety of methods have been proposed to train robust models directly on all data \cite{han2018masking, zhang2018generalized, yi2022learning}, while others try to identify label noise and reduce its negative effects with carefully designed strategies \cite{ortego2021multiobjective, karim2022unicon, li2022neighborhood, li2024regroup,yi2023classindependent}. Most previous works generally focus on closed-set noise, which assumes that noisy labels still belong to known classes. 

In more realistic scenarios, open-set noise from similar unknown classes may also occur in the training and inference stage \cite{wu2021ngc}, which poses significant challenges to LNL methods. This problem setup, called \textit{Learning with Open-world Noisy Data (LOND)}, is illustrated in the left subplot of Figure \ref{fig:setup}. For instance, in a cat-dog classification problem, images of cats and dogs may be mislabeled as each other, resulting in closed-set noise. Due to similar features, open-set noise also occurs in the training and inference stage, where the similar unknown classes of lion and wolf may be labeled as cat and dog. These open-set noises significantly degrade the performance of the existing LNL model \cite{yi2023classindependent}, as shown in the orange lines of the right subplot. Therefore, addressing the LOND problem requires the model to learn a classifier for known classes, and a detector for open-set noise, which is also known as an out-of-distribution (OOD) detector.

To address the LOND problem, we perform joint learning in dual representation space, i.e., the prototype space and the class-independent space, which is effective in dealing with closed-set and open-set noises (as shown in Figure \ref{fig:setup}). Specifically, to mitigate the model overfitting on the mixed noises, two networks are trained simultaneously, where the projection network uses learnable class prototypes to learn representations with shared semantics, and the One-Vs-All (OVA) network converts a multi-classification task into a multi-binary classification task by constructing a binary classifier for each class. Thus, the OVA network learns representations with unique semantics in the class-independent space, which reduces the inter-class competition in softmax and decouples the activations between clean and noisy labels. To enhance the detection of open-set samples, we introduce bi-level contrastive learning and consistency regularization in the two spaces. Subsequently, class-independent margin criteria are used to identify clean and noisy samples. For clean samples, the neighbor margin criterion aggregates OVA outputs from neighbor samples in the prototype space to ensure precise identification. 
On the other hand, open-set noise may have representations similar to closed-set samples, but its label can be considered as the negative class of all known classes. To measure the degree of open-set noise, the negative margin criterion is developed based on the closeness of negative probabilities in the class-independent space. For the remaining closed-set noise, we design a sample weight mechanism using the neighbor margin to measure the contribution to model learning. 
Our contributions are summarized as follows.

(1) We propose a novel joint learning framework with dual representation space using class-independent margin. Our method enhances performance on both classification for known classes and detection for open-set noise.

(2) To learn robust sample representation, bi-level contrastive learning and consistency regularization are introduced in dual representation space. Moreover, a margin-based sample identification mechanism is developed to effectively distinguish and use data with mixed noisy labels.

(3) We evaluate our method on multiple synthetic and real-world datasets, which demonstrates it achieves state-of-the-art (SOTA) performance.

\section{Related Work}
\subsection{Learning with Noisy Labels}
LNL methods aim to improve the model generalization from data with noisy labels and can be categorized into two main types based on how they use training data. \cite{huang2019o2unet}. The first type directly uses all data to train robust models, including robust model architectures \cite{han2018masking}, loss functions \cite{zhang2018generalized}, and regularization methods \cite{yi2022learning}. Since these methods treat the entire training data uniformly, their performance is limited at a high noise rate. The second type aims to identify noisy labels and reduce their negative effects using different criteria. The criteria commonly are based on the model output, such as cross-entropy loss \cite{li2019dividemix}, Jensen–Shannon divergence \cite{karim2022unicon}, regroup median loss \cite{li2024regroup}, logit margin \cite{zhang2024learning}. In addition to the outputs of a softmax classifier, representation neighbor information \cite{li2022neighborhood, ortego2021multiobjective} and the outputs of OVA network \cite{yi2023classindependent} are introduced for further performance improvement.

The above methods focus on addressing closed-set noise in the training set. Recent methods have studied the training set containing closed-set and open-set noises. For instance, to identify open-set noise, ILON \cite{wang2018iterative} and Rog \cite{lee2019robust} employ neighbor density and class-wise distribution estimation, respectively. SNCF \cite{albert2022embedding} performs spectral clustering to find representations of open-set noise. Jo-src \cite{yao2021josrc} and EvidentialMix \cite{sachdeva2021evidentialmix} exploit the consistency and uncertainty in model outputs to detect the mixed noises. However, these methods neglect the open-set noise during the inference. To the best of our knowledge, only NGC \cite{wu2021ngc} addresses the LOND problem using neighbor graphs. It obtains a softmax classifier and a prototype-based detector. However, the softmax classifier reduces the discriminability between clean and noisy labels due to the intersection of different class activations \cite{yi2023classindependent}. Moreover, the detector shows weak adaptability to the LOND problem, since its class prototype is simply updated by the average of class representations. On the other hand, our method learns a good classifier and detector from joint learning in the prototype and class-independent spaces.
\subsection{Out-of-distribution Detection}
OOD detection in classification tasks aims to detect samples from unknown classes and can be divided into two categories: training-based and post-hoc methods \cite{yang2024generalized}. The former requires retraining with extra or synthesized outlier data \cite{hendrycks2018deep, du2021vos} to enhance detection ability. In contrast, post-hoc methods can be directly performed on existing classifiers without retraining them, which maintains good classification performance \cite{yang2022openood}. These includes methods based on softmax \cite{guo2017calibration, hendrycks2017baseline}, logits \cite{hendrycks2022scaling, liu2020energybased, sun2021react, song2022rankfeat}, and distance \cite{bendale2016open, lee2018simple, sun2022outofdistribution}. Their training datasets are usually assumed to have clean labels, but this is hard to meet in real-world scenarios. A recent study \cite{humblot-renaux2024noisy} has analyzed the effect of label noise on post-hoc methods, but it still focuses on closed-set noise. Instead, our method learns to effectively detect samples with unknown classes from datasets containing closed-set and open-set noises.
\section{Method}\label{sec:method}
\begin{figure*}[t]
    \centering
    \includegraphics[width=\linewidth, trim = 0pt 5pt 0pt 0pt, clip]{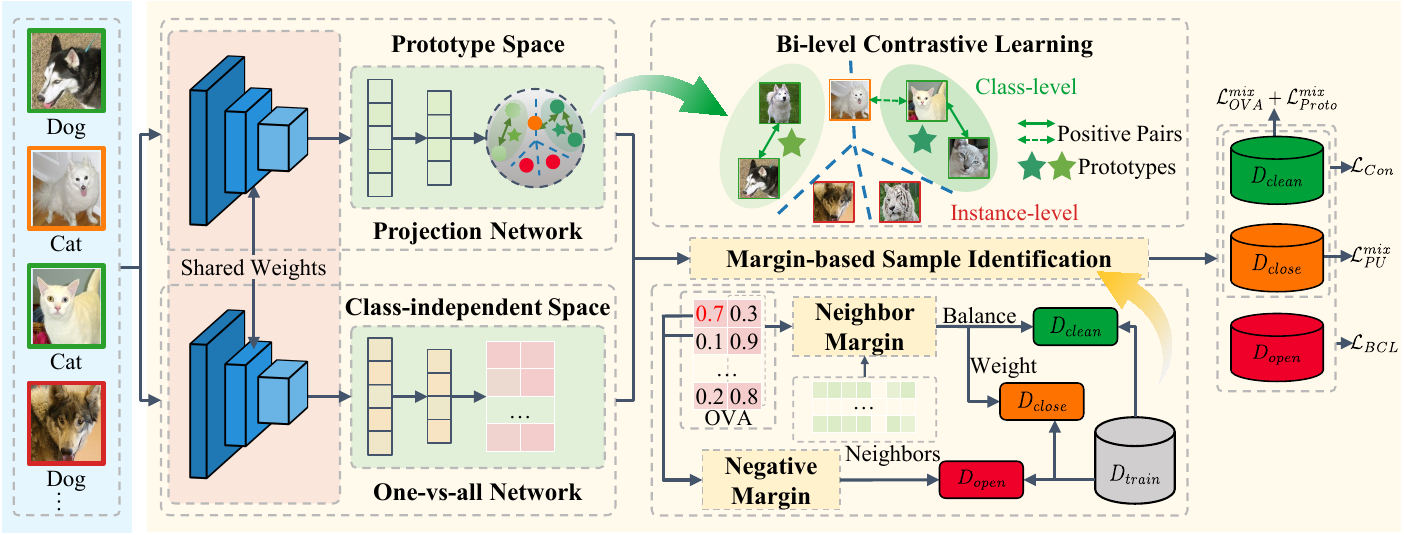}
    \caption{
        The overall framework of our proposed method. It uses projection and OVA networks to jointly learn in dual representation space, where bi-level contrastive learning and consistency regularization are introduced to enhance the detection of open-set noise. Then, class-independent margin criteria are used for sample identification. It uses the neighbor margin to select class-balanced clean samples $D_{clean}$, weighted closed-set noise $D_{close}$, and the negative margin to filter open-set noise $D_{open}$. Different losses are applied to these sample sets to obtain a classifier for known classes and a detector for open-set noise.
    }
    \label{fig:framework}
\end{figure*}
\subsection{Problem Statement}\label{sec:method_pro}
Formally, consider an image classification task. The training set is denoted as $D_{train}=\left\{(x_i,y_i)\right\}_{i=1}^{N}$, where $x_i$ is an input image and $y_{i}\in \mathcal{C} = \{1, \ldots, C\}$ is the given label. Since the labels of $D_{train}$ may not be correct, the true label of a sample $x_i$ is denoted as $y_i^{*}$. A clean sample has a correct label, i.e., $y_i=y_i^{*}$. Given $y_i \ne y_i^{*}$, the noisy sample falls into one of two categories: $y_i^{*}\in \mathcal{C}$ (closed-set noise) or $y_i^{*}\notin \mathcal{C}$ (open-set noise). The goal is to train a robust model that performs well on the test set $D_{test}=\{(x_i,y_i^{*})\}_{i=1}^{M}$. If $y_i^{*}\notin \mathcal{C}$ in the test set, the sample $x_i$ still belongs to the label space of the open-set noise in $D_{train}$ \cite{wu2021ngc}. 

\subsection{Overview}
To address the LOND problem, we propose a joint learning method in dual representation space, as illustrated in Figure \ref{fig:framework}. After input images are processed by a feature extractor with shared weights, the projection and networks are jointly learned in the prototype and class-independent spaces. To improve the detection ability of the model for open-set noise, we introduce bi-level contrastive learning on all data in the projection network and consistency regularization on closed-set data in the OVA network. Then, two class-independent margin criteria are developed for effective sample identification. Class-balanced clean samples are selected by measuring the consistency of the neighbor label, called the neighbor margin criterion. The negative margin is developed using the probabilities of negative classes in OVA outputs to filter the open-set noise. For the remaining closed-set noise, sample weights are generated by the neighbor margin. Finally, these sample sets are applied with different losses to obtain a classifier for known classes and a detector for open-set noise.

\subsection{Joint Learning in Dual Representation Space}
Existing LNL methods generally handle noisy labels using predictions from a softmax classifier \cite{yi2023classindependent}. However, the softmax has a competitive mechanism and generates dependent confidence scores among similar classes, making it sensitive to noisy labels. On the other hand, current class prototypes are updated directly by the average representations \cite{wu2021ngc}, which limits their adaptability to open-world scenarios. To address these problems, our method performs joint learning in dual representation space, i.e., the prototype and class-independent spaces, to effectively handle both closed-set and open-set noises. Specifically, the projection network constructs the prototype space based on learnable class prototypes by learning representations with shared semantics, thus improving its adaptability to open-set noise. The OVA network constructs the class-independent space by converting a multi-classification task into a multi-binary classification task, where each class is learned independently by a sub-classifier. This decouples the activations of different classes and learns representations with unique semantics, thus improving the discriminability between clean and noisy samples.

In the prototype space, the projection network learns class prototypes by gradient descent. Let the prototypes be the normalized vector set $P = \{P_c\}_{c=1}^{C}$. Denote the feature extractor as $G$, the projection head as $H$, and the sample representation as the embedding $z_i = H(G(x_i))$. The prototype loss for a sample $x_i$ is defined as
\begin{equation}
\setlength{\arraycolsep}{2pt}
\begin{array}[b]{ccc}
\mathcal{L}_{Proto}(x_i, y_i) = & -\underbrace{P_{y_i} \cdot z_i / \tau} & + \underbrace{\log \sum_{c=1}^{C} \exp(P_c \cdot z_i / \tau)} \\
& \text{tightness} & \text{contrastive}
\end{array},
\end{equation}
where $\tau$ denotes the temperature parameter and is simply set to 0.1 in all experiments. $\mathcal{L}_{Proto}$ considers intra-class tightness and inter-class separation to learn good representation.

In the class-independent space, let the OVA classifier be $F_{OVA}$. The output vector for the $c$-th class is $p_{OVA}^{c}(x_i) = F_{OVA}^{c}(G(x_{i})) = [p_{OVA}^{c} (z=0|x_i), p_{OVA}^{c}(z=1|x_i)]$, and $p_{OVA}^{c} (z=0|x_i) + p_{OVA}^{c}(z=1|x_i) = 1$. Here, $z=0$ and $z=1$ indicate the sample does not belong to and belongs to the class, respectively. Then, the OVA loss for a sample $x_i$ is defined as
\begin{equation}
\begin{split}
\mathcal{L}_{OVA}(x_i,y_i) = &- \log p_{OVA}^{y_{i}}(z=1|x_i) \\
& - \sum_{j=1,j\ne y_i}^{C} \log p_{OVA}^{y_{i}}(z=0|x_i).
\end{split}
\end{equation}

To further improve robustness to mixed noises, loss mixup is adopted, which provides better performance of OOD detection compared to label mixup \cite{pinto2022using}. The convex combination of $x_a$ and $x_b$ is defined as $x_i^{mix}= \lambda x_a + (1-\lambda) x_b$, where $\lambda \in [0,1] \sim Beta(\alpha, \alpha)$. The loss mixup is defined as
\begin{equation}
    \mathcal{L}_i^{mix}=\lambda \mathcal{L}_i (x_i^{mix}, y_a)+(1-\lambda)\mathcal{L}_i (x_i^{mix},y_b).
\end{equation}

\subsection{Open-set Robust Representation Learning}
Robust representation learning for noisy labels is generally achieved by contrastive learning \cite{yi2022learning} and consistency regularization \cite{li2019dividemix}. However, existing methods often ignore handling open-set noise. To alleviate this problem, we introduce robust representation learning for open-set noise, including bi-level contrastive learning in the prototype space and consistency regularization in the class-independent space. This enhances the ability of the model to detect open-set noise.

In the prototype space, contrastive learning is performed at the class and instance levels for the identified closed-set and open-set samples, respectively. On two-view data with strong and weak augmentations, the loss of the bi-level contrastive learning for a single sample $x_i$ is defined as follows:
\begin{equation}
    \mathcal{L}_{BCL}(z_i, y_i) = \frac{1}{1+|P(i)|} \mathcal{L}_{BCL,i},
\end{equation}
where $P(i)$ denotes the index set of views of other samples with shared labels $(y_{i}=y_{j})$ in a minibatch data $B$, and $|\cdot|$ indicates the cardinality of the set. $\mathcal{L}_{BCL,i}$ is defined as
\begin{equation}
\begin{split}
\mathcal{L}_{BCL,i} &= - \log \frac{exp(z_i \cdot z_i^{*} / \tau)}{\sum_{r=1, r \ne i}^{2|B|} exp(z_i \cdot z_r / \tau)} \\
&- \sum_{j\in P(i)} \log \frac{w(x_i) \cdot w(x_j) \cdot exp(z_i \cdot z_j / \tau)}{\sum_{r=1, r \ne i}^{2|B|} exp(z_i \cdot z_r / \tau)},
\end{split}
\end{equation}
where $z_{i}^{*}$ denotes another view of $x_i$, and $w(\cdot)$ is the sample weight derived from the neighbor margin. 

In the class-independent space, the loss of consistency regularization with respect to the output of the OVA network is defined as
\begin{equation}
\begin{split}
\mathcal{L}_{Con}(x_i, y_i) = \sum_{c=1}^{C} \sum_{j \in (0,1)} & || p_{OVA}^{c}(z=j|t_{s} (x_i)) \\
& - p_{OVA}^{c}(z=j|t_{w} (x_i)) ||^2_2, 
\end{split}
\end{equation}
where $t_{s}$ and $t_{w}$ denote strong and weak augmentations. 

\subsection{Margin-based Sample Identification}
Sample identification in LNL aims to distinguish clean and noisy samples based on estimated quality. In general, existing methods design estimation criteria based on the outputs of a softmax classifier, but they are hard to reduce inter-class competition in softmax \cite{yi2023classindependent}. Although the logits of a softmax classifier have been used to design margin criteria \cite{zhang2024learning}, they neglect open-set noise. To handle the mixed noises, we design the class-independent margin criteria in dual representation space, including the neighbor margin and the negative margin.

A clean sample has higher consistency between its given label and the neighbor label that is aggregated from the neighbor samples \cite{ortego2021multiobjective}. To obtain the neighbor label, previous methods aggregate the given labels or softmax probabilities of neighbors, which may not be effective at a high noise rate. Instead, we aggregate the OVA probability outputs in the class-independent space based on the representation neighbors. The probability of class $c$ of the neighbor label of sample $x_i$ is defined as follows:
\begin{equation}
q_{Neigh}^{c}(x_i) = \sum_{j=1}^k w_{ij}^{Neigh} p_{OVA}^{c}(x_i),
\end{equation}
where $k$ indicates the number of nearest neighbors, and $w_{ij}^{Neigh}$ denotes the normalized weight based on the distance between sample $x_i$ and neighbor sample $x_j$, which is defined as $w_{ij}^{Neigh} = exp(z_i \cdot z_j / \tau) / \sum_{j=1}^k exp(z_i \cdot z_j / \tau)$.

To measure the consistency between the given label and the neighbor label, the neighbor margin is defined as
\begin{equation}
\begin{split}
M_{Neigh}(x_i) &= q^{y_i}_{Neigh}(z=1|x_i) \\
&- \frac{1}{K} \sum_{j=1,j\ne y_i}^{K} q^j_{Neigh}(z=1|x_i),
\end{split}
\end{equation}
where $\sum_{j=1,j\ne y_i}^{K} q^j_{Neigh}(z=1|x_i)$ is the sum of the top-k probabilities. The larger the value of $M_{Neigh}^{i}$ ($x$ is omitted), the more likely it is to be the clean sample. To select clean samples, a naive method is to use a fixed threshold on the margin. However, since the margin is larger for easy classes and smaller for hard classes, this method can result in a class-imbalanced set of clean samples. To ensure class balance, the clean sample set for each class $c$ is defined as
\begin{equation}
D^{c}_{clean} = \left\{ (x_i, y_i): M_{Neigh}(x_i) \le \gamma_{c} \right\},
\end{equation}
where $\gamma_{c}$ is a class-wise dynamic threshold determined by the $\alpha_{ID}$-quantile of the class consistency degree that is the number of neighbor labels equal to given labels. 

Open-set noise may have similar features to closed-set samples with the same given label. Their representations with shared semantics make open-set noise difficult to identify. In fact, an open-set sample can be considered as not belonging to any known classes. In other words, it belongs to the negative class of all known classes. The OVA network estimates the probability that a sample does not belong to each class, called the negative probability. If the negative probability of a given label is close to that of other labels, the sample is identified as an open-set sample, belonging to the negative class of all known classes. Thus, the negative margin is defined as
\begin{equation}
M_{Neg}(x_i) = |p^{y_i}_{OVA}(z=0|x_i) - \max_{j\ne y_i} p^{j}_{OVA}(z=0|x_i) |.
\end{equation}
The smaller the value of $M_{Neg}$, the more likely it is an open-set noise sample. These samples are generally treated as a novel class and can be selected using a single threshold. Thus, the set of open-set noise is
\begin{equation}
\begin{split}
D_{open} = \{ (x_i, y_i) : &M_{Neg}(x_i) \le \gamma_{Neg}, \\
& (x_i, y_i) \notin D_{clean} \},
\end{split}
\end{equation}
where $\gamma_{Neg}$ is determined by the first $\alpha_{OOD}$ percentage of $M_{Neg}$ sorted in ascending order.

To effectively use the rest closed-set noise, the sample weights are defined as
\begin{equation}
w(x_i)= \begin{cases}
1, & \text {$x_i \in D_{clean}$} \\
0, &\text{$x_i \in D_{open}$} \\
(M_{Neigh}^{i} + 1) / (M_{Neigh}^{\max} + 1), &\text{$x_i \in D_{close}$}
\end{cases},
\end{equation}
where $M_{Neigh}^{\max}$ is the maximum margin and $D_{close} = D_{train} - D_{clean} - D_{open}$. 
In addition to bi-level contrastive learning, these sample weights are also used in the pseudo-label learning of class prototypes. After mixing closed-set samples as in \cite{li2019dividemix}, the pseudo-label loss for  prototypes of a closed-set sample $x_i$ is defined as
\begin{equation}
\begin{split}
\mathcal{L}_{PU}(x_i, y_i) = || p_i - Sharpen(\bar{y}_i, w(x_i), T) ||^2_2, \\
Sharpen(\bar{y}_i, w(x_i), T) = \left\{ \frac{\bar{y}_{i,c}^{w(x_i)/T}}{\sum_{j=1}^C \bar{y}_{i,j}^{w(x_i)/T}} \right\}_{c=1}^C,
\end{split}
\end{equation}
where $p_i$ is the softmax probability of the prototype logits $\left \{P_c \cdot z_i / \tau \right \}_{c=1}^C$, $\bar{y}_i$ is the average probability of the sample with strong and weak augmentations, and $T$ is the sharpening parameter that is set to 0.5 in this study.

\subsection{Training and Inference}
On the three identified sample subsets, the total loss of our method is
\begin{equation}
\begin{split}
\mathcal{L} &= \sum_{D_{clean}} \mathcal{L}_{OVA}^{mix} + \sum_{D_{clean}} \mathcal{L}_{Proto}^{mix} + \sum_{D_{close}} \mathcal{L}_{PU}^{mix} \\
&+ \lambda_{Con} \sum_{D_{clean} \cup D_{close}} \mathcal{L}_{Con} + \lambda_{BCL} \sum_{D_{train}} \mathcal{L}_{BCL}.
\end{split}
\end{equation}

During inference, the OVA output is used to evaluate the classification performance for known classes. For the OOD detection performance, the OVA output and the prototypes are combined to calculate the following OOD score:
\begin{equation}
s(x_i)=p_{OVA}^{y_{proto}}(z=0|x_i),
\end{equation}
where $y_{proto}$ is the label predicted by the class prototypes.

\section{Experiments}\label{sec:exp}
In this section, our method is compared with other SOTA methods on multiple datasets. Specifically, the effectiveness of our method is evaluated on closed-world, open-world, and real-world noisy data. The ablation study is also conducted to verify the importance of different modules of our method.

\begin{table*}[t]
\centering
\begin{tabular}{ccccccccc}
\toprule
\multirow{2}{*}{Methods} & \multirow{2}{*}{Publication} & \multicolumn{3}{c}{CIFAR100N}                    &           & \multicolumn{3}{c}{CIFAR80N}                     \\ \cmidrule{3-5} \cmidrule{7-9} 
                         &                              & Sym-20\%       & Sym-80\%       & Asym-40\%      &           & Sym-20\%       & Sym-80\%       & Asym-40\%      \\ \cmidrule{1-5} \cmidrule{7-9} 
Standard                 & -                            & 35.50          & 3.84           & 28.43          &           & 29.37          & 4.20           & 22.25          \\
Co-teaching              & NeurIPS 2018                 & 56.21          & 22.83          & 37.26          &           & 60.38          & 16.59          & 42.42          \\
Co-teaching+             & ICML 2019                    & 52.87          & 18.55          & 38.78          &           & 53.97          & 12.29          & 43.01          \\
DivideMix                & ICLR 2020                    & 57.76          & 28.98          & 43.75          &           & 57.47          & 21.18          & 37.47          \\
NGC\textsuperscript{\dag}                      & ICCV 2021                    & 60.95    & 40.10    & 45.50    &     & 64.67    & 37.12    & 47.83    \\
NCE                      & ECCV 2022                    & 54.58          & 35.23          & 49.90          &           & 58.53          & 39.34          & 56.40          \\
CIR+\textsuperscript{\dag}                     & AAAI 2023                    & 57.73    & 44.95    & 52.67    &           & 61.11    & 45.75    & 56.47    \\
NPN-hard\textsuperscript{\dag}                 & AAAI 2024                    & 65.27          & 36.88          & 60.11          &           & 66.07    & 35.38    & 64.09    \\
SED                      & ECCV 2024                    & 66.50          & 38.15          & 58.29          &           & 69.10          & 42.57          & 60.87          \\
Ours                     & -                            & \textbf{67.11} & \textbf{48.33} & \textbf{65.22} & \textbf{} & \textbf{69.61} & \textbf{49.30} & \textbf{67.27} \\ \bottomrule
\end{tabular}
\caption{Average accuracy (\%) on closed-world noisy data (CIFAR100N) and open-world noisy data (CIFAR80N) over the last ten epochs, where ``Sym" and ``Asym" denote the symmetric and asymmetric noise, respectively. }
\label{tab:cifar_acc}
\end{table*}

\begin{table*}[t]
\centering
\resizebox{\linewidth}{!}{ %< auto-adjusts font size to fill line
\begin{tabular}{c|ccc|cc|cccc|c|c}
\toprule
CIFAR80N  & MDS   & KNN   & OpenMax & MSP   & TempScaling & MLS   & EBO   & REACT & RankFeat & NGC   & Ours           \\ \midrule
Sym-20\%  & 48.02 & 44.47 & 60.93   & 63.92 & 64.43       & 63.73 & 63.45 & 62.90 & 56.54    & 67.97 & \textbf{75.97} \\
Sym-80\%  & 51.06 & 48.95 & 57.17   & 60.23 & 61.44       & 61.83 & 61.72 & 61.01 & 55.22    & 59.17 & \textbf{63.82} \\
Asym-40\% & 47.21 & 45.97 & 60.61   & 61.48 & 61.97       & 61.95 & 61.88 & 61.44 & 56.41    & 67.64 & \textbf{73.50} \\
Avg.      & 48.76 & 46.46 & 59.57   & 61.88 & 62.61       & 62.50 & 62.35 & 61.78 & 56.06    & 64.93 & \textbf{70.10} \\ \bottomrule
\end{tabular}}
\caption{AUROC (\%) comparison with SED combined with SOTA post-hoc methods and NGC, where ``Avg." denotes the average performance on three cases.}
\label{tab:cifar_auroc}
\end{table*}
\subsection{Experimental Setup}
\subsubsection{Datasets.}
The effectiveness of our method is evaluated on the CIFAR80N, CIFAR100N, Web-Aircraft, Web-Car, and Web-Bird datasets. Specifically, the CIFAR100 dataset \cite{krizhevsky2009learning}, containing 50,000 training images and 10,000 test images, is used as the base dataset. Synthetic noises are added to this dataset to generate the CIFAR80N and CIFAR100N datasets, following the settings in \cite{yao2021josrc}. They contain both symmetric and asymmetric types of noise with specified noise rates. In addition, the last 20 classes of CIFAR100 are added to the test set of CIFAR80N to validate the OOD detection performance of our method. To further test our method under more challenging scenarios, web datasets are used, including Web-Aircraft, Web-Car, and Web-Bird \cite{sun2021webly}. These datasets are collected via image search engines, inevitably resulting in unknown noise rates and complex noise types.

\subsubsection{Implementation Details.}
For experiments on the CIFAR datasets, a seven-layer CNN \cite{yao2021josrc} is used as the backbone network. It is trained using SGD with a momentum of 0.9, a weight decay of 0.0005, and an initial learning rate of 0.05 adjusted by cosine annealing. Both the batch size and the projection dimension are set to 128. Set $\alpha = 1$ in the Beta distribution, $K = 3$ for symmetric noise, and $K = 1$ for asymmetric noise. The network is trained for 300 epochs, including a 50-epoch warm-up phase. For OOD detection methods, the settings are consistent with those in \cite{yang2022openood}. For web datasets, ResNet50 pre-trained on ImageNet is adopted and trained using SGD consistent with the CIFAR experiments. This training uses a batch size of 64 and an initial learning rate of 0.005. The ResNet50 is trained for 120 epochs with a 10-epoch warm-up phase, where the prototype loss is added with a weight of 10 to fully use the pre-trained knowledge. Moreover, set $\alpha = 0.5$, $K = 1$, and $\alpha_{ID} = 0.5$.
\subsubsection{Baselines.}
On CIFAR100N and CIFAR80N, our method is compared with recent SOTA methods, including Co-teaching \cite{han2018coteaching}, Co-teaching+ \cite{yu2019how}, DivideMix \cite{li2019dividemix}, NGC, NCE \cite{li2022neighborhood}, CIR+ \cite{yi2023classindependent}, NPN-hard \cite{sheng2024adaptive}, and SED \cite{sheng2024foster}. The performance of the model trained using only cross-entropy (denoted as standard) is also shown. Most results of these methods are from SED, and \dag\ denotes the re-implemented performance using open-source code (except for NPN-hard on CIFAR100N). On the web datasets, we also compare the following SOTA methods: Jo-SRC, UNICON \cite{karim2022unicon}, and DISC \cite{li2023disc}. For OOD detection, SED is combined with multiple post-hoc methods. These include distance-based methods like MDS \cite{lee2018simple}, KNN \cite{sun2022outofdistribution}, and OpenMax \cite{bendale2016open}, and softmax-based methods like MSP \cite{hendrycks2017baseline} and TempScaling \cite{guo2017calibration}, logit-based methods like MLS \cite{hendrycks2022scaling}, EBO \cite{liu2020energybased}, REACT \cite{sun2021react}, and RankFeat \cite{song2022rankfeat}. Classification performance is measured by the top-1 accuracy metric and OOD detection is verified using the AUROC metric.

\subsection{Comparisons with State-of-the-art Methods}
\subsubsection{Evaluation on Closed-world and Open-world Noisy Data}
The results validated on the closed-world noisy data CIFAR100N are shown in Table \ref{tab:cifar_acc}. It is clearly evident that our method outperforms other methods. In the most challenging case (i.e., Sym-80\%), our method improves by at least 3.38\%. This indicates that our method can handle closed-set noise effectively.

On the open-world noisy data CIFAR80N, our method also outperforms other methods in both classification and OOD detection performance, as evidenced by Tables \ref{tab:cifar_acc} and \ref{tab:cifar_auroc}. 
For the classification performance in Table \ref{tab:cifar_acc}, our method still maintains an accuracy of 49.30\% at Sym-80\%. Toward more realistic asymmetric noise, our method improves the accuracy by at least 3.18\%. These results show that our method achieves superior classification performance even in the presence of high noise rates and complex noise types.
For the OOD detection performance in Table \ref{tab:cifar_auroc}, our method clearly achieves the best AUROC scores. Compared to other methods, our method increases at least 6.17\% on average. In all, our method can effectively identify open-set noise while robustly classifying known classes.

\subsubsection{Evaluation on Real-world Noisy Data.}
The comparison performance on real-world datasets is shown in Table \ref{tab:web_acc}. The results confirm that our method outperforms other SOTA methods on average. Specifically, our method achieves an accuracy of 89.92\% on Web-Aircraft, 80.95\% on Web-Bird, and 89.45\% on Web-Car, with an average improvement of at least 0.94\% over other methods. These results demonstrate that our method effectively handles real-world data with unknown noise rates and types. This can be attributed to the effective joint learning in dual representation space and the strategy of sample identification, which significantly reduces the negative effects of the mixed noises.
\begin{table*}[htbp]
\centering
\begin{tabular}{ccccccc}
\toprule
Methods      & Publication  & Backbone & Web-Aircraft & Web-Bird   & Web-Car    & Avg.       \\ \midrule
Standard     & -            & ResNet50 & 60.80        & 64.40      & 60.60      & 61.93      \\
Co-teaching  & NeurIPS 2018 & ResNet50 & 79.54        & 76.68      & 84.95      & 80.39      \\
Co-teaching+ & ICML 2019    & ResNet50 & 74.80        & 70.12      & 76.77      & 73.90      \\
DivideMix    & ICLR 2020    & ResNet50 & 82.48        & 74.40      & 84.27      & 80.38      \\
NGC          & ICCV 2021    & ResNet50 & 78.64        & 75.37      & 82.48      & 78.83      \\
Jo-SRC       & CVPR 2021    & ResNet50 & 82.73        & 81.22      & 88.13      & 84.03      \\
UNICON       & CVPR 2022    & ResNet50 & 85.18        & 81.20      & 88.15      & 84.84      \\
NCE          & ECCV 2022    & ResNet50 & 84.94        & 80.22      & 86.38      & 83.85      \\
DISC         & CVPR 2023    & ResNet50 & 85.27        & 81.08      & 88.31      & 84.89      \\
NPN-hard     & AAAI 2024    & ResNet50 & 86.02        & 80.91      & 88.26      & 85.06      \\
SED          & ECCV 2024    & ResNet50 & 86.62        & \textbf{82.00}      & 88.88      & 85.83      \\
Ours         & -            & ResNet50 & \textbf{89.92}   & 80.95 & \textbf{89.45} & \textbf{86.77} \\ \bottomrule
\end{tabular}
\caption{Accuracy (\%) comparison with SOTA methods on Web-Aircraft, Web-Bird, and Web-Car, where the results of existing methods are mainly copied from \cite{sheng2024foster}.}
\label{tab:web_acc}
\end{table*}
\begin{figure*}[t]
    \centering
    \begin{subfigure}{0.24\textwidth}
        \includegraphics[width=\linewidth]{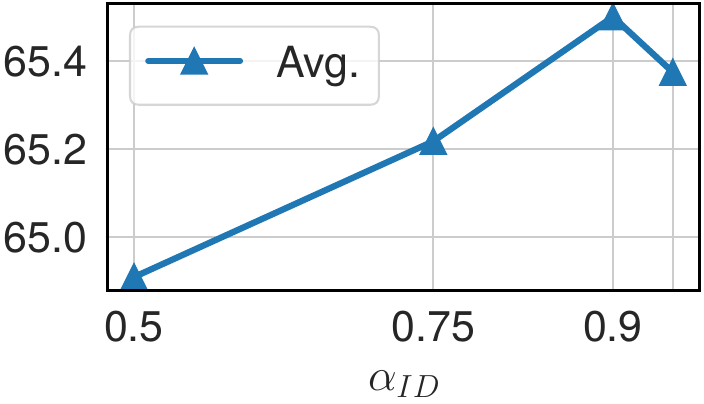}
        % \caption{First Image}
    \end{subfigure}\hfill
    \begin{subfigure}{0.24\textwidth}
        \includegraphics[width=\linewidth]{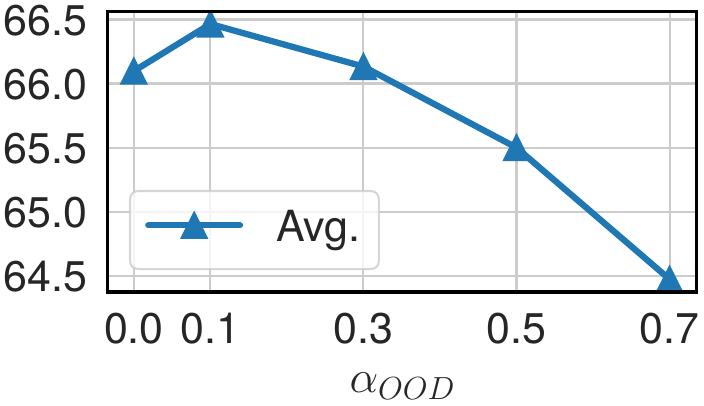}
        % \caption{Second Image}
    \end{subfigure}\hfill
    \begin{subfigure}{0.24\textwidth}
        \includegraphics[width=\linewidth]{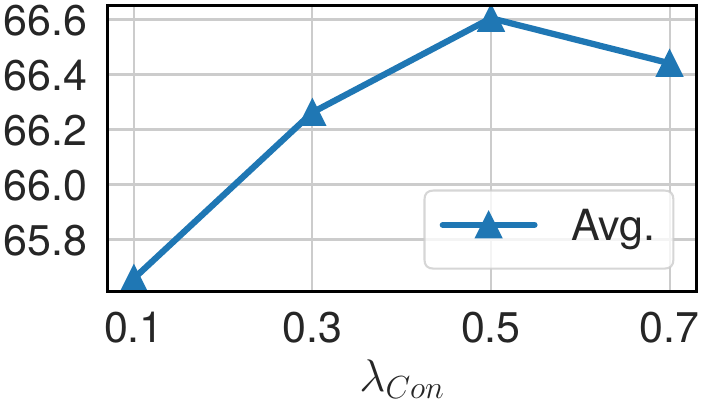}
        % \caption{Third Image}
    \end{subfigure}\hfill
    \begin{subfigure}{0.24\textwidth}
        \includegraphics[width=\linewidth]{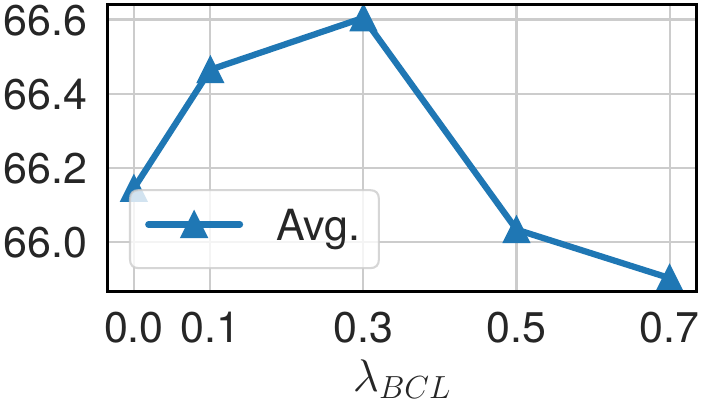}
        % \caption{Fourth Image}
    \end{subfigure}
    \caption{The sensitivity of hyper-parameters $\alpha_{ID}$, $\alpha_{OOD}$, $\lambda_{Con}$, and $\lambda_{BCL}$, where ``Avg." denotes the average of accuracy and AUROC of three cases (Sym-20\%, Sym-80\%, and Asym-40\%) on CIFAR80N.} \label{fig:hp_tuning}
\end{figure*}
\subsection{Ablation Study}
\subsubsection{Effects of Different Modules.}
An ablation study is conducted to investigate the contributions of different modules and losses used in the proposed method. Table \ref{tab:ablation} shows the classification and OOD detection performance at Sym-80\% and Asym-40\% on CIFAR80N. The average performance is also listed, which balances the classification performance and the OOD detection performance. 

It is evident that each module enhances the average metric. Our method employs joint learning in the prototype and class-independent spaces to mitigate mixed noise overfitting. It is worth noting that our warm-up model alone obtains 33.95\% accuracy at Sym-80\% and 55.66\% accuracy at Asym-40\%, which outperforms some SOTA methods in Table \ref{tab:cifar_acc}. In addition to warm-up, we use $\mathcal{L}_{OVA}$ and $\mathcal{L}_{Proto}$ on the clean samples as a baseline, where the samples are identified by the neighbor margin. This enhances the average performance by 8.47\% compared to (1). It suggests that learning in dual representation space could effectively handle closed-set and open-set noises. 
Moreover, $\mathcal{L}_{PU}$ is used for pseudo-label learning in the prototype space. This additional learning further enhances the performance on asymmetric noise. For example,  at Asym-40\%, the model with the loss of $\mathcal{L}_{PU}$ achieves a 1.00\% improvement compared to (2). This may be because the neighbor margin-weighted temperature smooths the noise prediction and reduces noise overfitting.
In addition, we apply $\mathcal{L}_{Con}$ on the closed-set samples in the class-independent space. Adding $\mathcal{L}_{Con}$ improves the average AUROC by 0.57\% on both types of noises. This may be attributed to that $\mathcal{L}_{Con}$ requires the model to predict consistently on closed-set samples with different perturbations.
On all training data, we further apply bi-level contrastive learning. Our proposed model improves the accuracy by 1.65\% at Sym-80\%. The reason for this may be that $\mathcal{L}_{BCL}$ can construct high-quality sample pairs from the instance and class levels for better representation learning.

\begin{table}
\centering
\setlength{\tabcolsep}{1mm}
{\fontsize{9pt}{\baselineskip}\selectfont
\begin{tabular}{lcccccc}
\toprule
\multicolumn{1}{c}{\multirow{2}{*}{Methods/Noise}} & \multicolumn{3}{c}{Sym-80\%}                     & \multicolumn{3}{c}{Asym-40\%}                     \\ \cmidrule{2-4} \cmidrule{5-7}
\multicolumn{1}{c}{}                               & ACC            & AUROC          & Avg.           & ACC            & AUROC          & Avg.           \\ \midrule
(1).Warmup                                         & 33.95          & 56.11          & 45.03          & 55.66          & 67.35          & 61.50          \\
(2).Baseline                                       & 46.95          & 62.68          & 54.82          & 66.15          & 71.16          & 68.65          \\
(3).(2)+$\mathcal{L}_{PU}$                     & 46.95          & 62.51          & 54.73          & 67.00          & 72.30          & 69.65          \\
(4).(3)+$\mathcal{L}_{Con}$                     & 47.65          & 63.03          & 55.34          & 66.40          & 72.92          & 69.66          \\
(5).(4)+$\mathcal{L}_{BCL}$                     & \textbf{49.30} & \textbf{63.82} & \textbf{56.56} & \textbf{67.27} & \textbf{73.50} & \textbf{70.39} \\ \bottomrule
\end{tabular}}
\caption{Ablation study of our method on CIFAR80N with 80\% symmetric and 40\% asymmetric noise rates.}
\label{tab:ablation}
\end{table}

\subsubsection{Sensitivity Analysis of Hyper-parameters.}
Our method introduces the parameter of $K$ in the neighbor margin, $\alpha_{ID}$ and $\alpha_{OOD}$ in the sample identification, and $\lambda_{Con}$ and $\lambda_{BCL}$ in the total loss function. Detailed analysis of $K$ is provided in the supplementary material, while the results for the other four hyper-parameters are shown in Figure \ref{fig:hp_tuning}. The performance is measured by the average metric of three cases (Sym-20\%, Asym-40\%, Sym-80\%) on CIFAR80N. 
The performance is enhanced when increasing the parameter of $\alpha_{ID}$ since more clear samples are selected. But after reaching the value of $0.9$, the performance is degraded dramatically. 
Conversely, the performance is decreased with a larger value of $\alpha_{OOD}$ because more potentially useful samples of known classes are filtered out.
For the trade-off parameters of losses, a moderate value of $\lambda_{Con}$ is preferred for balancing its weight with $\mathcal{L}_{OVA}$ to keep the classification performance.
Similarly, to achieve this balance, a smaller $\lambda_{BCL}$ should be taken. 
The method achieves the best balance between classification and OOD detection performance when $\alpha_{ID}=0.9$, $\alpha_{OOD}=0.1$, $\lambda_{Con}=0.5$, and $\lambda_{BCL}=0.3$.

\section{Conclusion}\label{sec:conclusion}
In this study, we propose the joint learning method with dual representation space to address the LOND problem. Specifically, projection and OVA networks are simultaneously trained to reduce the model overfitting on closed-set and open-set noises. To enhance the detection of open-set noise, our method introduces bi-level contrastive learning and consistency regularization. Moreover, class-independent margin criteria are designed to identify clean samples, closed-set noise, and open-set noise, and jointly minimize different losses on these sample subsets to benefit from the memorization effects. Our method outperforms other SOTA methods on the open-world noisy data. 

\section{Acknowledgments}
% TODO: acknowledgments
This work was supported in part by the National Natural Science Foundation of China (Grant Nos. 62476171, 62476172, 62076164, and 62206122), the Guangdong Basic and Applied Basic Research Foundation (Grant No. 2024A1515011367), the Guangdong Provincial Key Laboratory (Grant No. 2023B1212060076), and the Shenzhen Institute of Artificial Intelligence and Robotics for Society.

\bibliography{aaai25}
\appendix
\newpage

\twocolumn[
\begin{@twocolumnfalse}
\begin{center}
\textbf{\LARGE Appendix}
% \vspace{2em}
\end{center}
\end{@twocolumnfalse}
]
\input{appendix}

\end{document}

%% file: appendix.tex
\section{Method}
\subsection{Training Algorithm}

The pseudo-code of our method is listed in Algorithm \ref{algo}. 

\begin{algorithm}
\caption{Pseudo-code of Our Method}
\label{algo}
\begin{algorithmic}[1]
\REQUIRE Noisy training dataset $D_{train}$, model parameters $\theta$, warm-up epochs $E_{w}$, total epochs $E_{total}$.
\STATE Initialize the weights of all samples to 1.
\FOR{$epoch=1,2,\ldots,E_{w}$}
\FOR{$iteration=1,2,\ldots$}
\STATE Mixup the mini-batch data $B$;
\STATE Calculate $\mathcal{L}^{mix}_{OVA}$ using Eq. (2);
\STATE Calculate $\mathcal{L}_{BCL}$ using Eq. (5);
\STATE Update the model by $\mathcal{L}^{mix}_{OVA} + \mathcal{L}_{BCL}$;
\ENDFOR
\ENDFOR
\FOR{$epoch=E_{w}+1, \ldots, E_{total}$}
\STATE Select clean samples $D_{clean}$ using Eq. (9);
\STATE Filter open-set noise $D_{open}$ using Eq. (11);
\STATE Weight closed-set noise $D_{close}$ using Eq. (12);
\IF{$epoch == E_{w}+1$}
\STATE Initialize the class prototypes using the average embedding of each class in $D_{clean}$;
\ENDIF
\FOR{$iteration=1,2,\ldots$}
\STATE Mixup the mini-batch data $B$;
\STATE Calculate $\mathcal{L}^{mix}_{Proto}$ using Eq. (1);
\STATE Calculate $\mathcal{L}^{mix}_{OVA}$ using Eq. (2);
\STATE Calculate $\mathcal{L}^{mix}_{PU}$ using Eq. (13);
\STATE Calculate $\mathcal{L}_{Con}$ using Eq. (6);
\STATE Calculate $\mathcal{L}_{BCL}$ using Eq. (4);
\STATE Update the model using Eq. (14);
\ENDFOR
\ENDFOR

\end{algorithmic}
\end{algorithm}

\section{Experimental Details}
In this section, we first present the task settings considered in the experiments, followed by the details of the Web datasets. Then, the implementation details for different types of noisy data are introduced. Finally, we describe the evaluation metrics, including the classification of known classes, and detection of open-set noise, which is also known as out-of-distribution (OOD) detection.

\subsection{Task Settings}
Three different settings considered in our experiments are shown in Table~\ref{tab:setting}. Concretely, we evaluate the proposed method in the case of Learning with Open-world Noisy Data (LOND), Learning with Closed-world Noisy Data (LCND), and Learning with Real-world Noisy Data (LRND).
\begin{table}[htbp]
\small
\centering
\setlength{\tabcolsep}{1mm}
\begin{tabular}{c|ccc}
\toprule
\multirow{2}{*}{Tasks} & \multicolumn{2}{c|}{$D_{train}$}                               & $D_{test}$               \\ \cmidrule{2-4} 
                               & Closed-set Noise      & \multicolumn{1}{c|}{Open-set Noise} & Open-set Noise        \\ \midrule
LOND                           & \CheckmarkBold & \CheckmarkBold               & \CheckmarkBold \\
LCND                           & \CheckmarkBold & \XSolidBrush               & \XSolidBrush \\
LRND                           & \CheckmarkBold & \CheckmarkBold               & \XSolidBrush \\ \bottomrule
\end{tabular}
\caption{Task settings in the experiments.}
\label{tab:setting}
\end{table}

\subsection{Web Datasets}
We used the following Web datasets in our experiments:
\begin{itemize}
    \item \textbf{Web-Aircraft.} It has 13,503 training images and 3,333 test images from 100 different classes.
    \item \textbf{Web-Bird.} It includes 18,388 training images and 5,794 test images from 200 different classes.
    \item \textbf{Web-Car.} It contains 21,448 training images and 8,041 test images from 196 different classes.
\end{itemize}

The three fine-grained datasets are resized to $448 \times 448$ for training and testing. Notably, these datasets are collected via image search engines, inevitably resulting in unknown noise rates and complex noise types in their training sets. On the other hand, their test sets come from manually annotated datasets with the same fine-grained classes, i.e., FGVC-Aircraft \cite{maji2013fine}, CUB2002011 \cite{wah2011caltech}, and Stanford Cars \cite{krause2013object}, respectively.

\subsection{Implementation Details}

On the CIFAR datasets, a seven-layer CNN \cite{yao2021josrc} is used as the backbone network. It is trained using SGD with a momentum of 0.9, a weight decay of 0.0005, and an initial learning rate of 0.05 adjusted by cosine annealing. Both the batch size and the projection dimension are set to 128. In the mixup \cite{zhang2018mixup}, the $\alpha$ value in the Beta distribution is set to 1. For the neighbor margin, we set $K=3$ for symmetric noise and $K=1$ for asymmetric noise, with the number of neighbors $k=200$. The network is trained for 300 epochs, including a 50-epoch warm-up phase.

For OOD detection, we follow the settings described in \cite{yang2022openood} for the compared post-hoc methods. For these methods, we use a modified CIFAR100 validation set that retains samples from the same classes as CIFAR80N.

For the Web datasets, ResNet50 pre-trained on ImageNet is adopted and trained using SGD to keep the consistency with the CIFAR experiment. This training uses a batch size of 64 and an initial learning rate of 0.005. The ResNet50 is trained for 120 epochs with a 10-epoch warm-up phase, where the prototype loss is added with a weight of 10 to fully use the pre-trained knowledge. In the Beta distribution, the $\alpha$ value is set to 0.5. For the neighbor margin, we set the number of top-\textit{k} probabilities $K=1$, the number of neighbors $k=80$, and the clean sample selection ratio $\alpha_{ID} = 0.5$.

All experiments are conducted on the NVIDIA A100 GPU with 40GB of memory.

\subsection{Evaluation Metrics}
We use the following three criteria to verify the performance of our method.
\begin{itemize}
    \item \textbf{Classification Accuracy.} It is the top-1 classification accuracy averaged over all known classes in the test set.
    \item \textbf{AUROC.} AUROC is the Area Under the Receiver Operating Characteristic curve and can be calculated by the area under the curve formed by the True Positive Rate (TPR) and the False Positive Rate (FPR).
    \item \textbf{FPR95.} FPR95 is the False Positive Rate (FPR) when the recall reaches 95\%.
\end{itemize}

\section{Additional Experimental Results}
In this section, we first present additional results from the LOND setting. Subsequently, a sensitivity analysis is performed for the hyper-parameter in the neighbor margin. Then, we show the robustness of our proposed method in selecting clean samples. Finally, the learned feature representations are visualized to show the effectiveness of our method.

\subsection{Evaluation on Open-world Noisy Data}
For CIFAR80N, the FPR95 metric is used to evaluate the OOD detection performance. The result is shown at Table \ref{tab:cifar_fpr95}. These results demonstrate that our method can more accurately detect OOD samples compared to other methods. To further validate our method, multiple types of open-set noises are injected into the CIFAR100N dataset. Specifically, the test sets of TinyImagenet and Places-365 were randomly sampled to include a specified number of samples as open-set noise. In the remaining data, 10,000 samples were randomly selected for testing. Table \ref{tab:more_acc} shows that our method achieves superior classification performance for known classes. For OOD detection, Table \ref{tab:more} also demonstrates our method outperforms other methods on average.
\begin{table}[htbp]
\centering
\begin{tabular}{cccc}
\toprule
CIFAR80N  & MDS   & KNN   & Ours  \\ \midrule
Sym-20\%  & 94.50 & 93.48 & \textbf{86.79} \\
Sym-80\%  & 94.05 & 93.97 & \textbf{92.13} \\
Asym-40\% & 95.01 & 93.74 & \textbf{88.09} \\
Avg.      & 94.52 & 93.73 & \textbf{89.00} \\
\bottomrule
\end{tabular}
\caption{FPR95 (\%) comparison with SED combined with SOTA post-hoc methods, where ``Avg." denotes the average performance on three cases.}
\label{tab:cifar_fpr95}
\end{table}

\begin{table}[htbp]
\centering
\begin{tabular}{cccc}
\toprule
OOD dataset  & \# OOD & SED   & Ours           \\
\midrule
TinyImageNet & 10K    & 61.63 & \textbf{62.33} \\
             & 20K    & 60.94 & \textbf{63.25} \\
Places-365   & 10K    & 61.78 & \textbf{63.92} \\
             & 20K    & 61.20 & \textbf{62.58}  \\
\bottomrule
\end{tabular}
\caption{Accuracy (\%) comparison on CIFAR100N with multiple open-set noises, where closed-set noise is Sym-50\%.}
\label{tab:more_acc}
\end{table}

\begin{table*}[htbp]
\centering
\small
\setlength{\tabcolsep}{1mm}
\begin{tabular}{cc|ccc|cc|cccc|c}
\toprule
OOD Dataset  & \# OOD & MDS   & KNN   & OpenMax & MSP   & TempScaling & MLS            & EBO   & REACT & RankFeat & Ours           \\ \midrule
TinyImageNet & 10K    & 36.82 & 42.10 & 69.07   & 71.68 & 74.72       & 76.75          & 73.88 & 68.80 & 53.68    & \textbf{72.21} \\
             & 20K    & 29.63 & 29.44 & 71.52   & 74.07 & 77.04       & \textbf{78.62} & 73.82 & 69.91 & 49.80    & 73.34          \\
Places-365   & 10K    & 37.93 & 37.19 & 62.07   & 66.50 & 67.98       & 68.65          & 66.01 & 59.10 & 47.46    & \textbf{72.46} \\
             & 20K    & 32.58 & 26.41 & 65.10   & 70.85 & 72.80       & 72.80          & 67.91 & 62.24 & 52.92    & \textbf{73.29} \\
\bottomrule
\end{tabular}
\caption{AUROC (\%) comparison on CIFAR100N with multiple open-set noises, where closed-set noise is Sym-50\%.}
\label{tab:more}
\end{table*}

\subsection{Sensitivity Analysis of Hyper-parameter}
\begin{figure}[htbp]
    \centering
    \begin{subfigure}{0.48\columnwidth}
        \includegraphics[width=\linewidth]{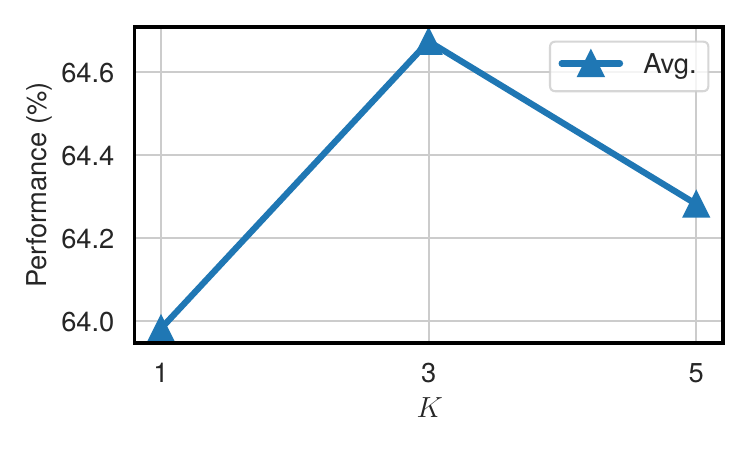}
        % \caption{First Image}
    \end{subfigure}\hfill
    \begin{subfigure}{0.48\columnwidth}
        \includegraphics[width=\linewidth]{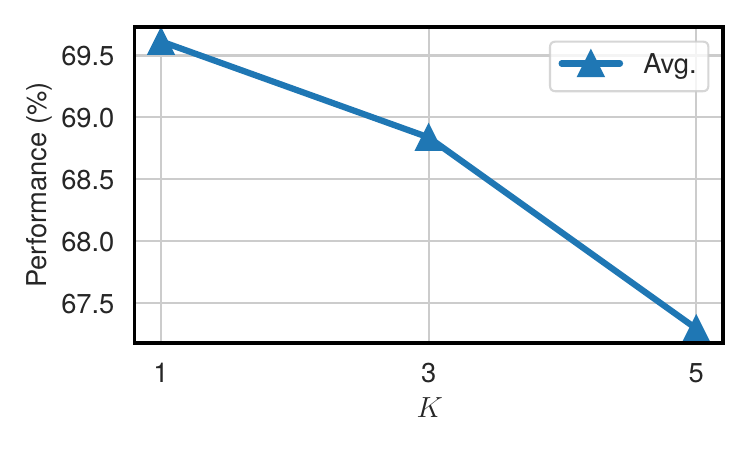}
        % \caption{Second Image}
    \end{subfigure}
    \caption{The sensitivity of $K$ on CIFAR80N with symmetric noise (left) and asymmetric noise (right).} \label{fig:hp_k}
\end{figure}

The effect of $K$ in the neighbor margin is investigated, and the results are shown in Figure \ref{fig:hp_k}. We vary the value of $K$ from 1 to 5 and validate it on symmetric and asymmetric noises. In symmetric noise, clean labels are randomly flipped to any class, thus a moderate value of $K$ can balance information from similar or dissimilar classes and achieve better performance. However, in asymmetric noise, clean labels tend to flip to similar classes. Therefore, the best model performance is achieved when $K=1$ since only the most similar class needs to be distinguished for selecting clean samples. Given the Web datasets contain more asymmetric noise, a smaller $K$ is more suitable. Therefore, we take $K=3$ for symmetric noise and $K=1$ for both asymmetric noise and the Web datasets.

\subsection{Robustness of Sample Selection}
\begin{figure}[htbp]
    \centering
    \begin{subfigure}{0.48\columnwidth}
        \includegraphics[width=\linewidth]{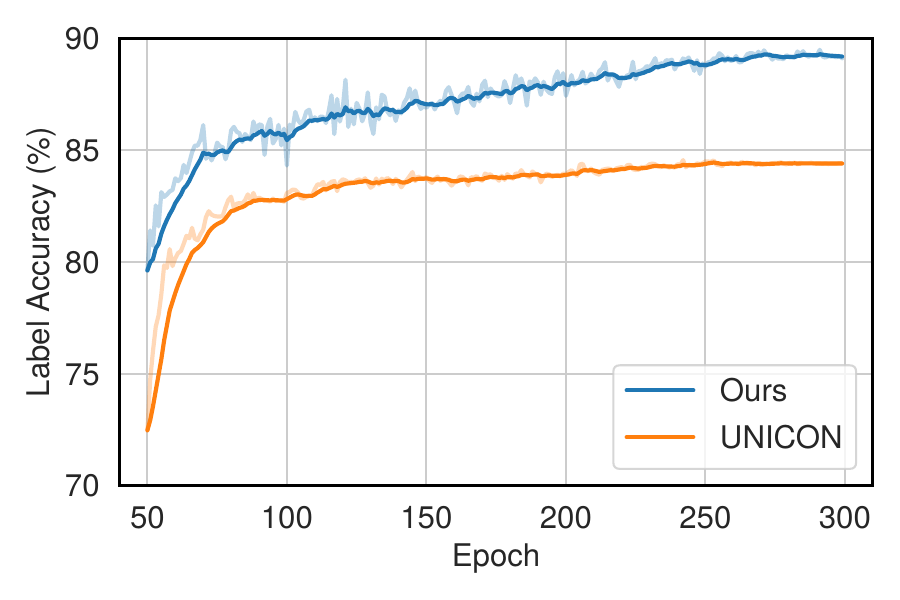}
        % \caption{First Image}
    \end{subfigure}\hfill
    \begin{subfigure}{0.48\columnwidth}
        \includegraphics[width=\linewidth]{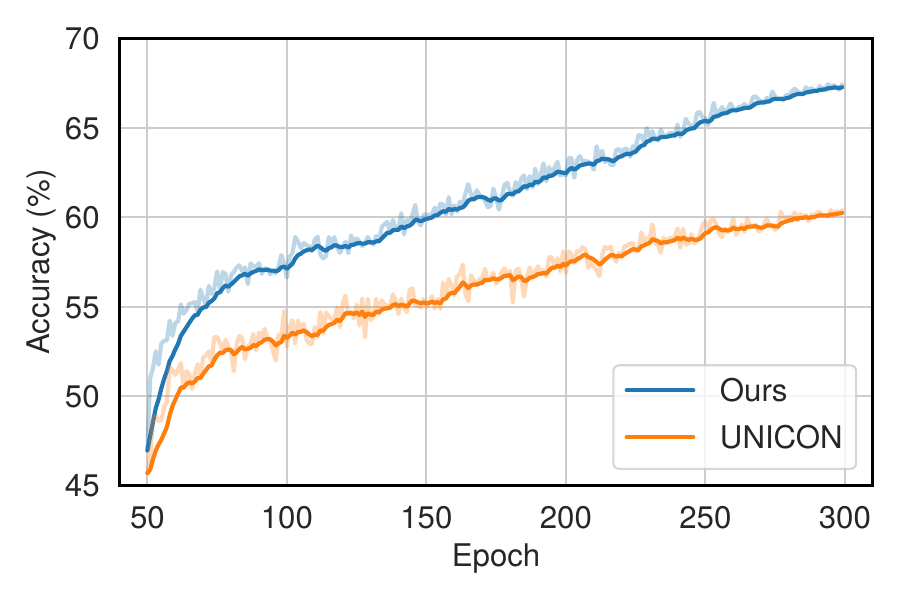}
        % \caption{Second Image}
    \end{subfigure}
    \caption{Comparison of our method and UNICON on CIFAR80N with 40\% asymmetric noise rate.} \label{fig:robust}
\end{figure}

To evaluate the robustness in selecting clean samples, the neighbor margin strategy is replaced with UNICON \cite{karim2022unicon} for comparison. The results are shown in Figure \ref{fig:robust}. After a 50-epoch warm-up phase, our method demonstrates better label accuracy and model performance. This may be attributed to that joint learning in dual representation space is helpful to enhance the discriminative ability between clean and noisy labels.

\subsection{Visualization of the Learned Feature Representation}
\begin{figure*}[htbp!] 
    \centering
    \begin{subfigure}{.32\textwidth}
        \includegraphics[width=\linewidth]{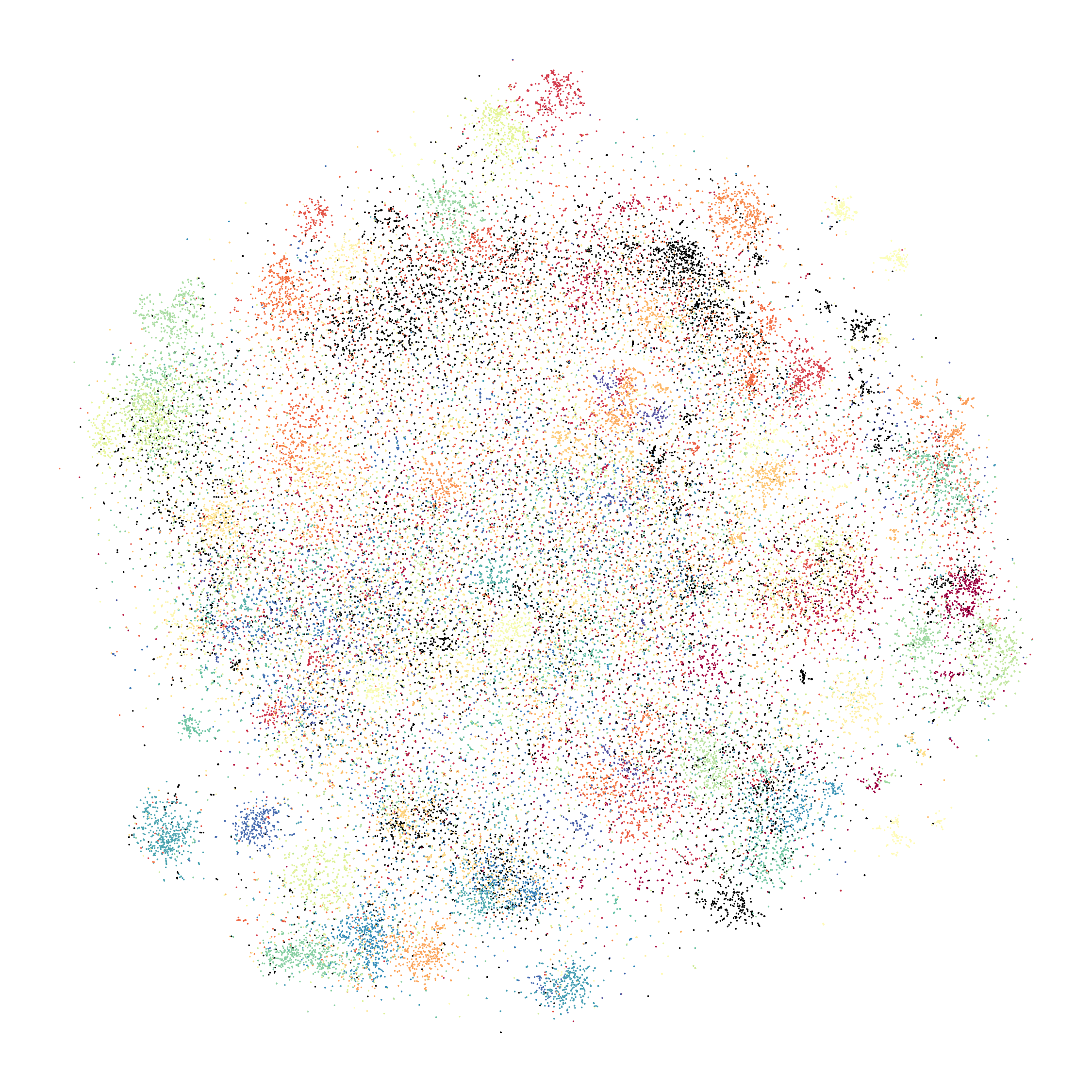} 
        \caption{SED}
    \end{subfigure}\hfill
    \begin{subfigure}{.32\textwidth}
        \includegraphics[width=\linewidth]{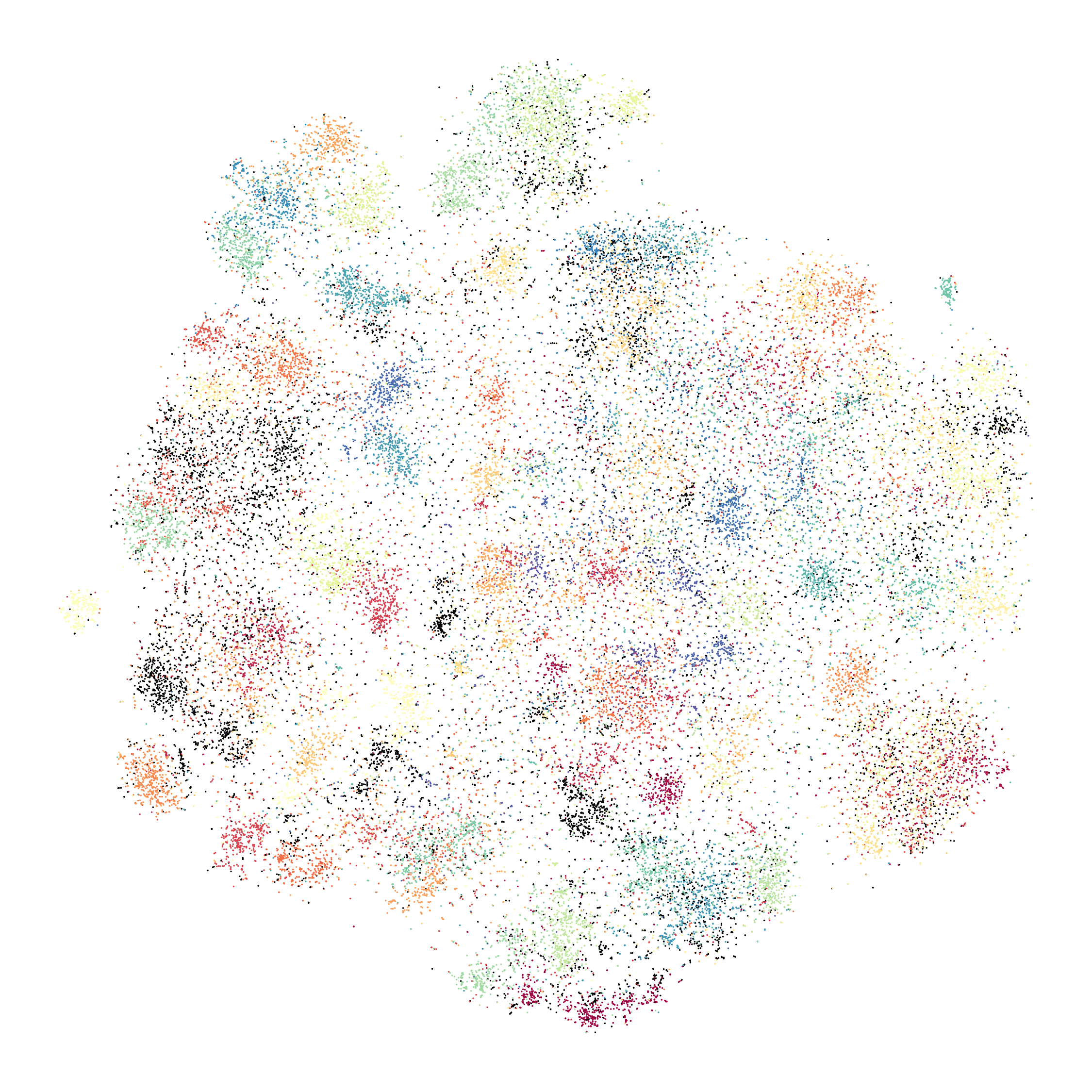}
        \caption{NGC}
    \end{subfigure}\hfill
    \begin{subfigure}{.32\textwidth}
        \includegraphics[width=\linewidth]{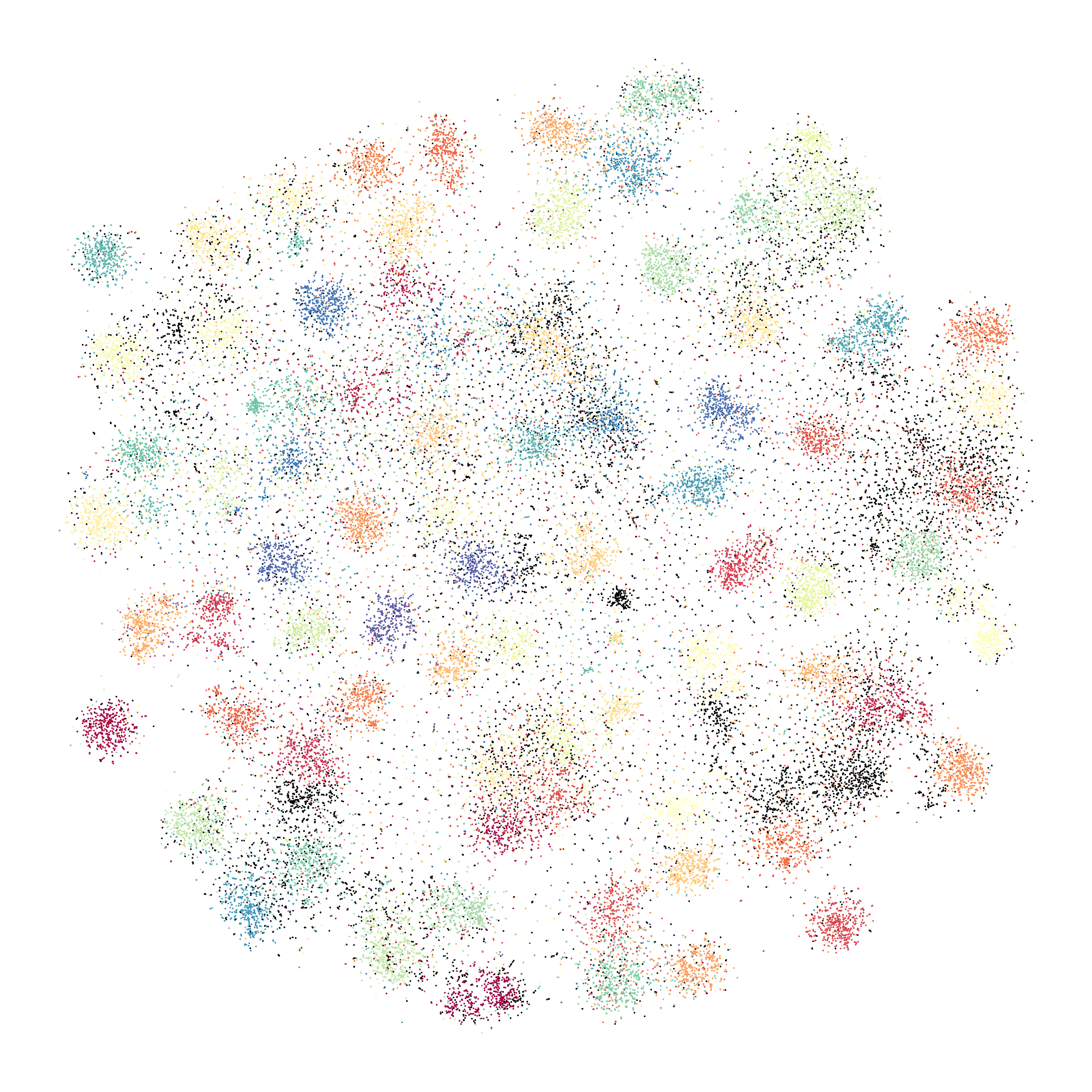}
        \caption{Ours}
    \end{subfigure}
    \caption{Visualization of learned feature representations on CIFAR80N at Sym-20\%, where open-set noise is colored in black.} \label{fig:vis}
\end{figure*}

To verify the effectiveness of joint learning in dual representation space, the learned sample representations are visualized using tSNE \cite{maaten2008visualizing}. Figure \ref{fig:vis} shows the results of different methods on CIFAR80N with 20\% symmetric noise rate. Our method obtains better separation and compactness for known classes and open-set noise. These results demonstrate that our method can learn robust representations to enhance the classification performance for known classes and the detection ability for open-set noise.
